\pdfoutput=1

\documentclass[11pt]{article}

\usepackage[]{ACL2023}

\usepackage{times}
\usepackage{latexsym}

\usepackage[T1]{fontenc}

\usepackage[utf8]{inputenc}

\usepackage{microtype}

\usepackage{inconsolata}

\usepackage{microtype}
\usepackage{amsfonts}
\usepackage{booktabs}
\usepackage{multirow}
\usepackage{amstext}
\usepackage{bm}
\usepackage{cite} 
\usepackage{amsmath}
\usepackage{graphicx}
\usepackage{subfigure}
\usepackage{xcolor}
\usepackage{CJKutf8}
\usepackage{diagbox}
\usepackage{cleveref}
\usepackage{url}

\title{A Multi-Modal Context Reasoning Approach \\ for Conditional Inference on Joint Textual and Visual Clues}

\author{Yunxin Li$^{1}$\thanks{~~$^{\dagger}$Corresponding author.}, Baotian Hu$^{1\dagger}$, Xinyu Chen$^{1}$, Yuxin Ding$^{1}$, Lin Ma$^{2}$, Min Zhang$^{1}$\\
$^{1}$Harbin Institute of Technology, Shenzhen, China, $^{2}$Meituan, Beijing\\
\texttt{\{hubaotian, yxding, zhangmin2021\}@hit.edu.cn}
\\
\texttt{\{liyunxin987, chenxinyuhitsz\}@163.com},
\texttt{forest.linma@gmail.com}
}

\begin{document}
\maketitle

\begin{abstract}

Conditional inference on joint textual and visual clues is a multi-modal reasoning task that textual clues provide prior permutation or external knowledge, which are complementary with visual content and pivotal to deducing the correct option. Previous methods utilizing pretrained vision-language models~(VLMs) have achieved impressive performances, yet they show a lack of multimodal context reasoning capability, especially for text-modal information. To address this issue, we propose a \textbf{M}ulti-m\textbf{od}al \textbf{C}ontext  \textbf{R}easoning approach, named \textit{ModCR}. Compared to VLMs performing reasoning via cross modal semantic alignment, it regards the given textual abstract semantic and objective image information as the pre-context information and embeds them into the language model to perform context reasoning. Different from recent vision-aided language models used in natural language processing, ModCR incorporates the multi-view semantic alignment information between language and vision by introducing the learnable alignment prefix between image and text in the pretrained language model. This makes the language model well-suitable for such multi-modal reasoning scenario on joint textual and visual clues. We conduct extensive experiments on two corresponding data sets and experimental results show significantly improved performance (exact gain by 4.8\% on PMR test set) compared to previous strong baselines. Code Link: \url{https://github.com/YunxinLi/Multimodal-Context-Reasoning}.

\end{abstract}

\section{Introduction}

\begin{figure}
    \centering
    \includegraphics[width=0.49\textwidth]{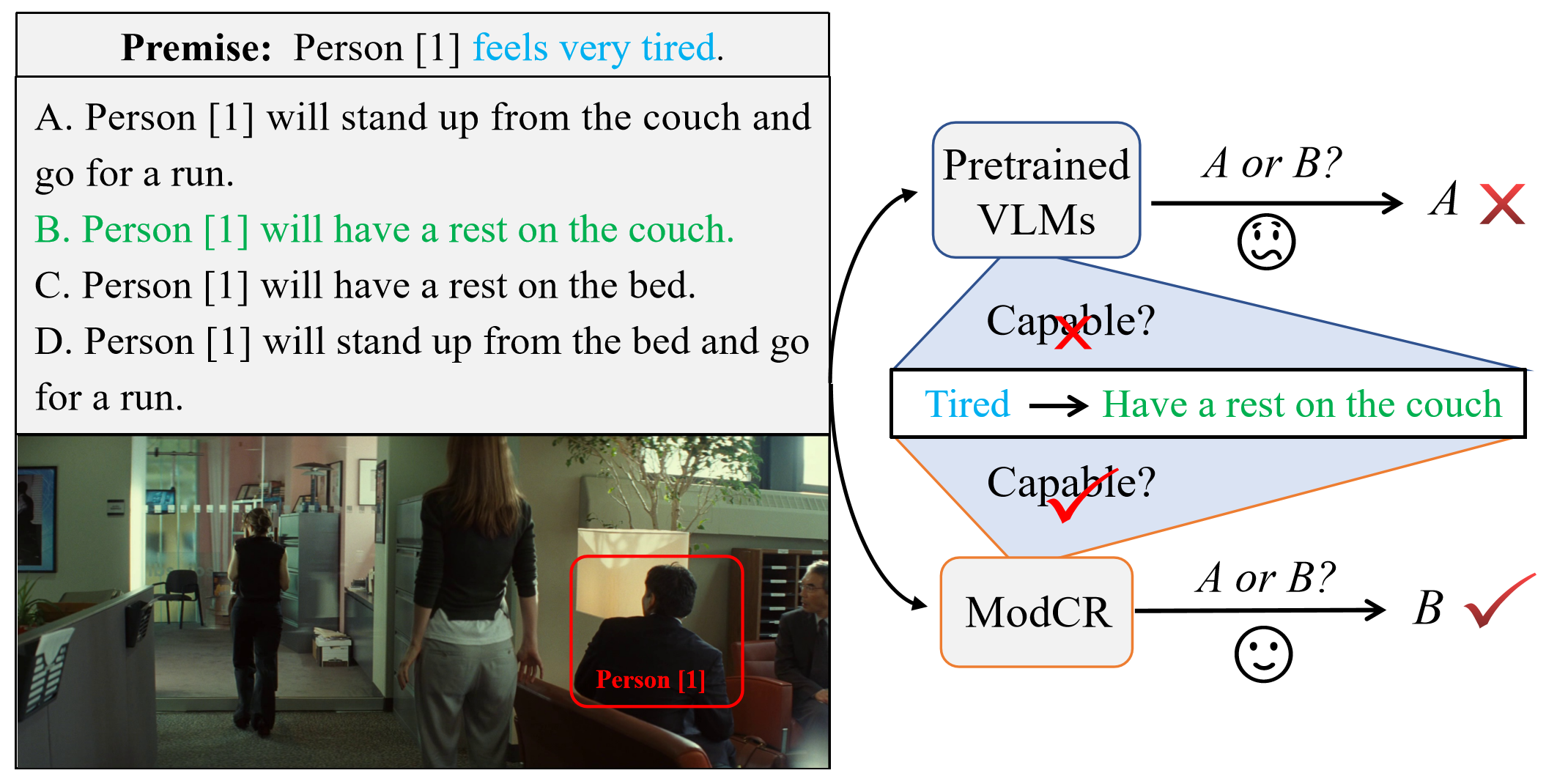}
    \caption{A case from the PMR~\citep{dong2022premise} data set, where the correct option is answer \textcolor{green}{B}. The blue-color words represent the pivotal textual clue to infer the correctness of answers A and B.}
    \label{fig:caseintro}
\end{figure}

Cross modal reasoning is a hot research topic both in natural language processing and computer vision communities. Most cross modal reasoning tasks, such as Visual Question Answering~\citep{antol2015vqa,wu2017visual_survey, shah2019kvqa, yusuf2022analysis}, Visual Dialog~\citep{zhang2022reasoning, chen2022utc}, Visual Entailment,~\citep{xie2019visual, do2020snli} and Visual Commonsense Reasoning~\citep{dataset_vcr, ye2021case, li2022representation}, concentrate on the visual reasoning scenario that relies primarily on image information. The given text~(or question) is highly attached to the image and lacks prior permutation, e.g., the common question \textit{``Why is person 4 pointing to person 1"} shown in VCR~\citep{dataset_vcr} data set. For another practical cross modal reasoning scenario~\citep{dong2022premise}, the textual modality often provides prior permutation or complementary information with the source image, such as the commonsense knowledge, and the personalities, feelings, or relationships of persons, as the premise shown in Figure~\ref{fig:caseintro}. In this paper, we focus on such conditional inference on joint textual and visual clues, where the specific task form is to select the correct option from the candidate set according to the given textual premise and image. 

Previous methods~\citep{chen2020uniter, imagescode, li2020oscar, dong2022premise, ofa} usually input the concatenated sequence of textual premise, image, and candidate answer into powerful pretrained vision-language models (VLMs) and employ a task-specific classifier to infer the result with attention to the joint representation obtained from VLMs. Although these methods work well for reasoning based mainly on visual clues, they suffer from one major shortcoming: the reasoning process does not fully utilize the abstract semantic information of given premise text to perform in-context reasoning. As the case shown in Figure~\ref{fig:caseintro}, pretrained VLMs know \textit{``person [1] sits on the couch, not the bed''} from the image, yet struggle to effectively infer that the person will \textit{``have a rest on the couch"} according to  \textit{``feels very tired''} presented in the premise. It may be attributed to that pretrained VLMs mostly map different modalities into a unified space~\citep{long2022vision} and perform cross modal semantic alignment and fusion. They neglect the in-context learning based on the given multi-modal semantics of language and vision during pertaining, like next sentence prediction. 
Fortunately, pretrained language models~(PLMs) such as BERT~\citep{devlin-etal-2019-bert}, RoBERTa~\citep{liu2019roberta}, BART~\citep{lewis-etal-2020-bart}, and GPT3~\citep{brown2020language}, are powerfully capable of in-context learning and have achieved successful performance on natural language inference and open-ended text generation. PLMs can infer the next-step intent according to the given abstract text information compared to pretrained VLMs. Hence, we propose a simple and effective Multi-modal In-Context Reasoning approach named \textit{ModCR} for this multi-modal reasoning task, taking advantages of VLMs and PLMs.


Specifically, ModCR employs a pretrained visual encoder equipped with a vision mapping network to obtain the image representation and convert it into the learnable visual prefix. The visual prefix and textual premise are regarded as two types of pre-context. They will be fed to the in-context reasoner, i.e., language model, to infer the correctness of answer. Considering the semantic gap between visual prefix and text in the language model, we first utilize a multi-grained vision-language semantic alignmenter to gain the multi-view alignment representation between image and text. Afterwards, we devise an alignment mapping network to capture the pivotal alignment information and convert it into the learnable cross-modal alignment prefix. Finally, we fed the two prefixes, premise, and answer into the language model to perform cross modal reasoning in the instruction template-based slot-filling method. In this way, ModCR bridges the semantic gap between visual content and text in the language model through introducing the cross-modal alignment prefix. It makes use of the abstract semantic of premise and objective image information via the self-attention mechanism in PLMs.


To verify the effectiveness of ModCR, we conduct extensive experiments on two cross modal reasoning data sets: PMR~\citep{dong2022premise} and VCR~\citep{dataset_vcr}. The experimental results show that the proposed method significantly outperforms previous strong baselines. The ablation and case studies indicate that ModCR is capable of in-context reasoning based on multi-modal information.

Our contributions can be summarised as follows:

\begin{itemize}
    \item We propose a multi-modal in-context reasoning framework for conditional inference on joint textual and visual clues, utilizing the in-context learning capability of PLMs.
    \item To the best of our knowledge, we are the first to introduce the multi-view alignment information between vision and language into the language model to perform cross modal reasoning, bridging the semantic gap between vision and language in PLMs.
    \item Experimental results show that ModCR achieves state-of-the-art performance on two corresponding data sets. It significantly outperforms previous vision-aided language models and pretrained VLMs-based approaches. 
    
\end{itemize}

\begin{figure*}[t]
\centering
    \includegraphics[width=0.90\textwidth, height=0.52\textwidth]{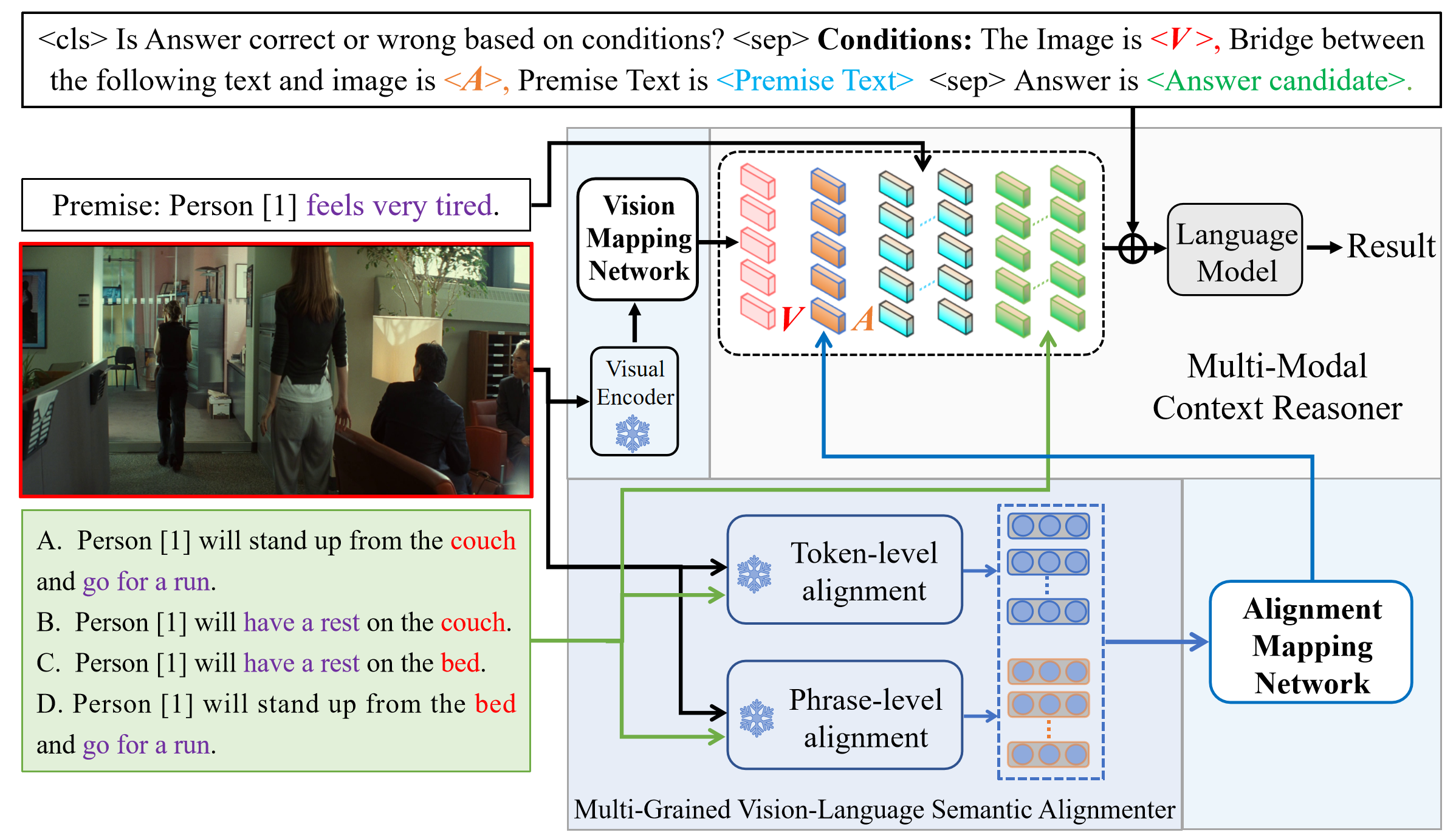}
    \caption{The overall workflow of ModCR. The top part presents the slot-filling instruction template used in the multi-modal in-context reasoner. The purple words show the relevant content between the premise and answers. The red words in answers are related to the image information. ``V'' and ``A'' indicate the vectors of visual and cross-modal alignment prefixes, respectively.   
    }
    \label{fig:model}
\end{figure*}


\section{Related Works}

 \textbf{Pretrained VLMs for Cross Modal Reasoning.}
Cross modal reasoning ~\citep{chen2021cross,long2022vision} is a challenging task that requires a cross modal understanding of images and texts with relational reasoning to infer the correct option. Vision-language models are thus proposed to represent, align, and fuse the image and text information and perform task-specific reasoning such as Visual Question Answering~\citep{antol2015vqa,wu2017visual_survey, shah2019kvqa, yusuf2022analysis,gao2022transform}, Visual Dialog~\citep{zhang2022reasoning, chen2022utc,lin-byrne-2022-retrieval} or Storytelling~\citep{huang2016visual, yu2021transitional}, Visual Entailment,~\citep{xie2019visual, do2020snli}, Visual Commonsense Reasoning~\citep{dataset_vcr, ye2021case, li2022representation}. 
Over the past few years, significant performance has been made for developing vision-language models, owing to the Transformer~\citep{attention} architecture and large-scale multi-modal web data~\citep{bugliarello-etal-2021-multimodal, lin2021m6}. These pretraind VLMs could be divided into single-stream~\citep{wang2021simvlm, li2020unimo} and double-stream~\citep{CLIP, jia2021scaling, lu2022cots} types according to multi-modal information interaction methods. Our work explores how to expand and ameliorate pretrained VLMs to conditional inference on joint textual and visual clues. 

\noindent\textbf{Vision-aided Language Models.}
Images can provide explicit and diverse visual information to improve the imaginative representation of language. Recent works show that vision-aided language models have achieved promising performance on natural language understanding~\citep{lu2022imagination} and open-ended text generation tasks~\citep{zhu2022visualize} such as text completion~\citep{zellers-etal-2019-hellaswag}, story generation~\citep{fan-etal-2018-hierarchical}, and concept-to-text~\citep{barzilay2005collective}. Some works~\citep{shi-etal-2019-visually, lu2022imagination} proposed to retrieve images corresponding to texts from the image corpus and use visual knowledge to improve the performance on the downstream tasks. Recently, some researchers~\citep{long-etal-2021-generative, yang2021open, zhu2022visualize} proposed to utilize the powerful text-to-image technical to obtain the imagination representation of language and infuse them into the language model via the prefix-tuning~\citep{li-liang-2021-prefix} way. In this paper, we also compared the visual prefix-based prompt learning methods~\citep{liang-etal-2022-modular, jin-etal-2022-good, tsimpoukelli2021multimodal}, which has been verified to improve the performance of pretrained language models.  
\section{Methodology}
\subsection{Overview}
ModICR focuses on infusing the given multi-modal information: premise, image, and answer, into the language model to make conditional inferences based on textual and visual clues. The overview of ModICR is illustrated in Figure~\ref{fig:model}. Specifically,
given the premise $P=({p_1,...,p_M})$, image $I$ and answer candidates~$A=({a_1,...,a_Y})$, where $p_i$, $a_i$ indicate the $i$ th token of premise and the $i$ th answer in the candidate set respectively, we first use the visual encoder to obtain the image representation, which is projected into the visual prefix to provide the objective environment information. Considering a semantic gap between visual prefixes and text when the language model performs context learning, we devise an alignment mapping network based on a multi-grained vision-language semantic alignmenter to gain the cross-modal alignment prefix. Finally, the two-type prefixes, premise text, and answer candidate are fed to the language model via the instruction learning way to perform multi-modal context reasoning.


\subsection{Base Model}
\label{basic}

Previous methods~\citep{dong2022premise, chen2020uniter, yu2021ernie} adopt the pretrained vision-language model to obtain joint representation of text and image during inferring. Similarly, we utilize the pretrained single-stream bidirectional encoder Oscar~\citep{li2020oscar} as the backbone of the visual encoder and multi-grained vision-language semantic alignmenter. In this case, the image feature is first extracted by the widely-used tool FasterRCNN~\citep{fasterrcnn} and fed into the visual encoder and alignmenter. Oscar mainly make the token-level semantic alignment between image and text. Hence, following~\citet{chunks}, we pretrain Oscar-based chunk-aware semantic interactor on the Flickr30k Entities~\citep{csi} data set to perform the phrase-level semantic alignment between text and image.

\subsection{Mapping Networks}
\label{mn}
We denote the obtained sequence representation of the image and the text aligned with the image features to $\mathbf{H}_{I} = (\mathbf{h}_{Ig}, \mathbf{h}_{I_1}, ..., \mathbf{h}_{I_O})$, $\mathbf{H}_{ta} = (\mathbf{h}_{tag}, \mathbf{h}_{ta_1}, ..., \mathbf{h}_{ta_N})$, and $\mathbf{H}_{pa} = (\mathbf{h}_{pag}, \mathbf{h}_{pa_1}, ..., \mathbf{h}_{pa_N})$, respectively, where $\mathbf{h}_{I_i}$ indicates the output hidden state of $i$ th image region (obtained by FasterRCNN). $\mathbf{h}_{ta_i}$ or $\mathbf{h}_{pa_i}$ represents the token-level or phrase-level aligned representation of $i$ th token in answer text. $N$ is the token length of answer. Similarly, $\mathbf{h}_{Ig}$, $\mathbf{h}_{tag}$, and $\mathbf{h}_{pag}$ show the global representations of image, token-level and phrase-level alignment information, respectively. However, the obtained visual and alignment embedding vectors may lie in a representation space different from the language model (used in the multi-modal context reasoner) due to the discrepancy across models. To alleviate this gap, we adopt the feature mapping network~\citep{mokady2021clipcap} to project them into the corresponding learnable prefixes.

\noindent\textbf{Vision Mapping Network (VMN).}  
As the top blue part shown in Figure~\ref{fig:model}, we use the visual encoder to encode the image and employ a vision mapping network to project image representation $\mathbf{H}_{I}$ into the sequence of visual prefix $\mathbf{V}=(v_1,...,v_l)$ with the mixed length $l$. $v_i$ represents the $i$ th visual embedding. The workflow is
\begin{equation}
    v_1,...,v_l = \text{VMN}(\mathbf{h}_{Ig}).
\label{eq1}
\end{equation}
For VMN, we adopt a two-layer perceptron with a ReLU activation function. It could be pretrained on large-scale image-text pairs for projecting visual features into the visual prefix that has the same space distribution as word embedding in LMs. 

\noindent\textbf{Alignment Mapping Network~(AMN).} 
It is capable of capturing the multi-view semantic alignment information of image-text pair and converting it into the cross-modal alignment prefix. Such prefix can bridge the semantic gap between visual prefix and text in the language model, enhancing the interactive understanding of image-text information. Specifically, we first apply a two-layer transformer to capture the pivotal multi-view alignment information lied in $\mathbf{H}_{ta}$ and $\mathbf{H}_{pa}$. The specific calculation process of the first layer is as follows:
\begin{equation}
    \begin{array}{c}
        \mathbf{h}_{dr} = \mathbf{W}^{dr}([\mathbf{h}_{tag}, \mathbf{h}_{pag}]) + \mathbf{b}^{dr}, \vspace{1.0ex}\\
        \mathbf{h}_{cr}=\text{cross}(\mathbf{h}_{dr}, [ \mathbf{h}_{ta_1}, ..., \mathbf{h}_{ta_N},  \mathbf{h}_{pa_1}, ..., \mathbf{h}_{pa_N}]), \vspace{1.0ex}\\
        \mathbf{h}_{ag}^{1} = \text{MLP}(\mathbf{h}_{cr}),
    \end{array}
\label{eq2}
\end{equation}
where $\mathbf{W}^{dr}$ and $\mathbf{b}^{dr}$ are learnable parameters. $cross$ represents the cross-attention calculation process. $\text{[},\text{]}$ shows the concatenate computation. After doing the same two-layer calculation, we obtain the pivotal alignment representation $\mathbf{h}_{ag}$. Secondly, we project it into the cross-modal alignment prefix via a similar calculation process as the vision mapping network (Eq.~\ref{eq1}). Finally, we gain an alignment prefix representation $\mathbf{A}=(a_1,...,a_m)$, where $a_i$ indicates the $i$ th alignment embedding and $m$ is the length of prefix. By doing so, AMN could capture the pivotal semantic alignment information and project them into the learnable prefix vectors in the word embedding space.

\subsection{Multi-Modal Context Reasoner}
\label{mmir}

After obtaining two types of the prefix, we infuse them into an context reasoner to conduct cross modal reasoning, where we adopt the pretrained language model RoBERTa~\citep{liu2019roberta} as the context reasoner. We utilize the widely used instruction-learning method to incorporate the whole context encoding information. Specifically,
we fill visual prefix, alignment prefix, premise and answer candidate in a pre-defined instruction template, \textit{``<cls> Is Answer correct or wrong based on conditions? <sep> Conditions: The Image is <$\mathbf{V}$>, Bridge between the following text and image is <$\mathbf{A}$>, Premise Text is <Premise Text> <sep> Answer is <Answer candidate>. ''}. These special symbols, <$\mathbf{V}$>, <$\mathbf{A}$>, \textit{<Premise Text>}, and \textit{<Answer candidate>}, will be replaced by the obtained prefix vectors $\mathbf{V}$ and $\mathbf{A}$, and word embedding representations of premise and answer in turn. The sequence representation is fed into the context reasoner to infer the final result. This way, we can utilize the context learning capability of pretrained language model to tackle the multi-modal reasoning problem.
We obtain the inferring result of each answer candidate by applying a two-layer perceptron with the ReLU activation function on the output hidden state $\mathbf{h}_{cls}$ of the top layer in RoBERTa. The whole training objective of ModICR can be defined as 
\begin{equation}
    \mathcal{L}_{f} = -\sum_{i=1}^{4}\mathbf{log}{P}_{i}(x_i=q),
\label{eq3}
\end{equation}
where ${x}_{i}$ is the output probability on $i$ th answer candidate and $q$ is the label. 

\subsection{Training and Inference}

To make Eq.~\ref{eq2} in the alignment mapping network capture pivotal multi-view alignment information, we will first train it about one epoch for alleviating the cold start problem leading to the collapse of the network. Concretely, we use a linear function to project $\mathbf{h}_{ag}$ into the confidence score and employ the cross entropy loss to optimize it locally with the golden label $q$. The training process is regarded as $\mathcal{L}_{1}$. Thus, the whole training process could be defined as
$$ \mathcal{L} =\left\{
\begin{array}{rcl}
\mathcal{L}_{1},       &      & {steps < N_{whole}},\\
\mathcal{L}_{f},      &      & {steps > N_{whole}},\\
\end{array} \right. $$
where $steps$ shows the optimization step during training and $N_{whole}$ represents the start of the whole training. 

For inference, we input each answer candidate with premise and image into ModICR to obtain the confidence score and adopt the maximum one as the final result.

\label{training}

\section{Experiment}

\subsection{Data sets}
Conditional inference on joint textual and visual clues is a task that the text provides the prior permutation or the complementary information (external knowledge) with the image. There are few data sets that meet the above requirement in the community. To verify the effectiveness of the proposed model, we first adopt the high-quality human-constructed PMR~\citep{dong2022premise} data set, which contains 12,080 training samples, 1,538 validation samples and 1,742 testing samples. Textual premises pass the human cross-check annotation and contain six categories: relationship, personality, mood, and so on. In addition, we also reorganized a corresponding large-scale data set according to the VCR data set~\citep{dataset_vcr}. We combine the given correct rationale and question as the textual premise and reform the original task into inferring the answer based on the new premise and image, i.e., QR$\rightarrow$A. This way, the rationale could provide external knowledge information different from the source image. We set the original validation as the test set and selected some training samples as the validation set. Finally, the samples are divided into 210k training/2,923 validating/ 26,534 testing.

\subsection{Baselines}
We compare the proposed method to pretrained LMs and VLMs as follows:

\textbf{BERT}~\citep{devlin-etal-2019-bert} and \textbf{RoBERTa}~\citep{liu2019roberta} are both the transformer-based large language model, having achieved impressive performance on many natural language understanding tasks. We fine-tune them with only access to the textual premise.

\textbf{VL-BERT}~\citep{vilbert} is a dual-stream pretrained cross-modal model. It adopts the BERT architecture, and the visual feature are concatenated with text embedding. 

\textbf{ERNIE-VL}~\citep{yu2021ernie} is a single-stream fusion encoder. It utilizes the structured knowledge obtained from scene graphs to learn joint representations of vision and language.

\textbf{UNITER}~\citep{chen2020uniter} also expands the BERT architecture to incorporate visual information and power heterogeneous downstream vision-language tasks with joint multi-modal embeddings.


\textbf{Oscar}~\citep{li2020oscar} is also a single-stream fusion encoder that uses object tags detected in images as anchor points to ease the learning of alignments significantly.

\textbf{OFA}~\citep{ofa} is a sequence-sequence cross-modal learning framework that unifies a diverse set of cross-modal and unimodal tasks, including visual grounding, image captioning, image classification, language modelling, etc.

\textbf{MVPTR}~\citep{mvptr} is a pretrained cross model that introduces the multi-level semantic alignment of vision-language to facilitate representation learning synergistically.

\textbf{CALeC}~\citep{chunks} is a unified prediction and generation model for some vision-language tasks, which introduces the chunk-aware semantic interactor to improve the semantic alignment representation and uses the lexical constraint technical to promote the quality of generation. 

\textbf{PromptFuse}~\citep{liang-etal-2022-modular} is a prompt-based learning method to infuse visual information into the language model. It randomly initializes two learnable vectors as the alignment prefix to improve the space representation projection of image and text and bridge the semantic gap between the visual prefix and text. 

\begin{table}[t]
\renewcommand\arraystretch{1.10}
\begin{center}
    \scalebox{0.82}{
    \begin{tabular}{c c c c }
        \toprule
        \textbf{Method} $\downarrow$ \textbf{Types} $\rightarrow$ & Validation & Testing \\
        \midrule
        BERT-B~\citep{devlin-etal-2019-bert} & - & 65.2\\
        VL-BERT-B~\citep{vilbert} & - & 75.4\\
        ERNIE-VL-B~\citep{yu2021ernie} & - & 79.0\\
        UNITER-B~\citep{chen2020uniter} & - & 77.4\\
        Oscar-B~\citep{li2020oscar} & 77.7 & 76.1\\
        RoBERTa-L~\citep{liu2019roberta} & 77.3& 75.0\\
        PromptFuse~\citep{liang-etal-2022-modular} & 77.4 & 76.5\\
        VL-BERT-L~\citep{vilbert} & - & 79.3\\
        ERNIE-VL-L~\citep{yu2021ernie} & - & \underline{79.9}\\
        UNITER-L~\citep{chen2020uniter} & - & 77.0\\
        OFA-L~\citep{ofa} &79.9 & 79.1\\
        MVPTR~\citep{mvptr} & 79.5 & 78.9\\
        CALeC~\citep{chunks} & \underline{80.1} & 78.7\\
        \midrule
        ModCR~(frozen VLMs) & 85.0 & 84.3\\
        ModCR~(fine-tune VLMs) & \textbf{85.8} & \textbf{84.7}\\
        \bottomrule
    \end{tabular}}
    \caption{\label{tab:pmr_result} Model performance (accuracy) on the PMR data set. The results of BERT, VL-BERT, ERNIE-VL, and UNITER are reported by \citet{dong2022premise}. For baselines, ``-B'' and ``-L'' indicate the base and large version, respectively. The underscore and bold indicate the second highest value and best performance (same as following tables). ``frozen VLMs'' and ``fine-tune VLMs'' represent whether the parameters of the visual encoder and multi-grained vision-language alignmenter are involved in training.}
\end{center}
\end{table}

\begin{table}[t]
\renewcommand\arraystretch{1.10}
\begin{center}
    \scalebox{0.76}{
    \begin{tabular}{c c c c c}
        \toprule
        \textbf{Method} $\downarrow$ \textbf{Types} $\rightarrow$ & AT $\uparrow$& D1 $\downarrow$& AF $\downarrow$& D2$\downarrow$\\
        \midrule
        BERT-B~\citep{devlin-etal-2019-bert} & 65.2 & 19.8 & 19.6 & 4.5\\
        Oscar-B~\citep{li2020oscar} & 76.1 & 10.2 & 12.1 & 1.7\\
        RoBERTa-L~\citep{liu2019roberta} & 75.0 & 17.7 & 6.1 & 1.2\\
        PromptFuse~\citep{liang-etal-2022-modular} & 76.5 & 16.5 & 5.8 & 1.2\\
        ERNIE-VL-L~\citep{yu2021ernie} & 79.9 & 10.7 & 8.2 & 1.2\\
        OFA-L~\citep{ofa} & 79.1 & 9.7 & 9.9 & 1.3\\
        MVPTR~\citep{mvptr} & 78.9 & \textbf{7.5} &  11.8 & 1.8\\
        CALeC~\citep{chunks} & 78.7 & 8.6 & 10.9 & 1.8\\
        \midrule
        ModCR~(frozen VLMs)  & 84.3 & 9.2& \textbf{5.6} & 0.9\\
        ModCR~(fine-tune VLMs) & \textbf{84.7}& 7.8 & 6.8 & \textbf{0.7} \\
        \bottomrule
    \end{tabular}}
    \caption{\label{tab:reasoning_result} Detailed performance of models on the test set of PMR. The results of BERT and ERNIE-VL are reported by \citet{dong2022premise}. AT, D1, AF, D2 represent the Action True and Image True, Action True yet Image False, Action False yet Image True, Action False and Image False, respectively. ``Action True or False'' indicate the answer whether meets the premise. Similarly, ``Image True or False'' show the answer whether meets the image information.}
\end{center}
\end{table}

\subsection{Implementation Details}
 We use the Adam~\citep{kingma2014adam} optimizer to train the above models on 2 A100 GPUs with a base learning rate of 2e-5, a batch size of 32, and a dropout rate of 0.1. For each sample, we set the maximum number of visual regions extracted by FasterRCNN to 10. We set $N_{whole}$ to 1 epoch and adopt the pre-trained parameters of the base version of Oscar to initialize the multi-grained vision-language semantic alignmenter. While training the chunk-level semantic interactor on the Flickr30k Entities data set, we follow the parameter settings presented in~\citet{chunks} and train it for about ten epochs. We adopt the Roberta$_{large}$ to initialize the multi-modal context reasoner. The visual and cross-modal alignment prefix lengths are both set to 5. All methods performed on the two data sets employ the validation set to select the best-performing model.

\subsection{Main Results}

\begin{table}[t]
\renewcommand\arraystretch{1.10}
\begin{center}
    \scalebox{0.82}{
    \begin{tabular}{c c c c }
        \toprule
        \textbf{Method} $\downarrow$ \textbf{Types} $\rightarrow$ & Validation & Testing \\
        \midrule
        Oscar-B~\citep{li2020oscar} & 87.3 & 86.0\\
        RoBERTa-L~\citep{liu2019roberta} & \underline{92.7}& \underline{91.8}\\
        OFA-L~\citep{ofa} & 90.3  & 89.4\\
        MVPTR~\citep{mvptr} & 84.2 & 85.3\\
        CALeC~\citep{chunks} & 90.8 & 90.5\\
        \midrule
        ModCR~(frozen VLMs) & 94.5 & 93.6\\
        ModCR~(fine-tune VLMs) & \textbf{94.7} & \textbf{94.0}\\
        \bottomrule
    \end{tabular}}
    \caption{\label{tab:vcr_result} Model performance (accuracy) on the validation and testing sets of VCR (QR$\rightarrow$A) data set.}
\end{center}
\end{table}

\textbf{Overall Performance.}
We report the performance of models on PMR and VCR (QR$\rightarrow$A) data sets, which are shown in Tables~\ref{tab:pmr_result} and~\ref{tab:vcr_result}. From the whole experimental results, we observe that the proposed method significantly outperforms previously strong baselines such as gain by 5.7\%, 4.8\% on the validation and testing of the PMR data set compared to CALeC and ERNIE-VL-L. According to the performance of BERT-B and RoBERTa (only text input), we know that the premise can provide vital information to infer the correct option. The performance is further improved when combined with visual content and cross-modal semantic alignment prefix for inference, e.g., ModCR (frozen VLMs) vs. RoBERTa: 84.3 vs. 75.0, PromptFuse vs. RoBERTa: 76.5 vs. 75.0. For model performances on VCR (QR-A), however, we observe that the pretrained VLMs have worse performance compared to RoBERTa-L, which displays that VLMs do not make good use of the abstract semantics of the premise for contextual reasoning. ModCR that takes the RoBERTa-L as the main backbone surpasses pretrained VLMs and LMs on two data sets, which suggests that our method effectively utilizes the semantic information of different modalities while performing reasoning. 

\begin{table}[t]
\renewcommand\arraystretch{1.10}
\begin{center}
    \scalebox{0.82}{
    \begin{tabular}{c c c c }
        \toprule
        \textbf{Method} $\downarrow$ \textbf{Types} $\rightarrow$ & Validation & Testing \\
        \midrule
        CALeC~\citep{chunks} & 80.1 & 78.7\\
        RoBERTa-L~\citep{liu2019roberta} & 77.3& 75.0\\
        PromptFuse (LV=1, LA=2) & 77.4 & 76.5\\
        \midrule
        ModCR~(LV=1, LA=0) & 78.1 & 76.0 \\
        ModCR~(LV=3, LA=0) & 78.2 & 77.8 \\
        ModCR~(LV=5, LA=0) & 77.3 &  76.8\\
        \midrule
        ModCR~(LV=3, LA=1) & 84.9 & 83.5 \\
        ModCR~(LV=3, LA=5) & \textbf{85.8} & 83.9\\
        ModCR~(LV=3, LA=7) & 85.3 & 84.1\\
        \midrule
        ModCR~(LV=1, LA=1) & 84.0 & 82.3\\
        ModCR~(LV=3, LA=3) & 84.8 & 83.8\\
        ModCR~(LV=5, LA=5) & 85.0 & \textbf{84.3}\\
        ModCR~(LV=7, LA=7) & 85.1 & 82.8\\
        ModCR~(LV=10, LA=10) & 79.7 & 79.3\\
        \bottomrule
    \end{tabular}}
    \caption{\label{tab:module_result} The experimental results of ModCR with different prefix length on the PMR data set. We frozen the parameters of VLMs for all ModCR variants. ``LV'' and ``LA'' indicate the lengths of visual and alignment prefix respectively, where ``=0'' represents that the corresponding mapping network is removed.}
\end{center}
\end{table}

\noindent\textbf{Is Context Reasoning Capability Improved?} We present the detailed performances of models on the test set of PMR to check the ability of models to infer different types of answer candidates, which contain AT, D1, AF, and D2, as shown in Table~\ref{tab:reasoning_result}. The reported results indicate that RoBERTa better uses the abstract semantic information of premise to infer the correctness of the following action compared to VLMs, e.g., RoBERTa without visual information has the lowest error rate across all baselines in action recognition (AT). In addition, we also find that although the ability of recently proposed VLMs to reason with abstract textual clues has been improved, there is still a particular gap compared to LMs, e.g., AT performance: OFA-L (8.2) vs. RoBERTa (6.0). When employing the language model RoBERTa as the reasoner and infusing the visual information in it, we observe that the overall accuracy of the model is further improved. However, the previous vision-infusing method has a low utilization rate of visual information (D1: 16.5 for PromptFuse). As the bottom two lines shown in Table~\ref{tab:reasoning_result}, ModCR, which utilizes the multi-view text-image semantic alignment information, maintains the abstract reasoning ability based on premise and also substantially improves the utilization rate of image information.

Through the above analysis, we can obtain that it is necessary to introduce vision-language semantic alignment information for vision-aided language models. Furthermore, there is still a large room for improvement in the contextual reasoning capability of the pretrained VLMs.


\begin{table}[t]
\renewcommand\arraystretch{1.10}
\begin{center}
    \scalebox{0.80}{
    \begin{tabular}{c c c|c c}
        \toprule
        MappNet & RoBERTa & VLM & Validation & Testing\\
        \hline
         $\checkmark$ & $\times$& $\times$ & 85.7 & 85.8\\
         $\checkmark$ & $\checkmark$& $\times$ & 94.5 & 93.6\\
         $\checkmark$ & $\checkmark$& $\checkmark$ & 94.7 & 94.0\\
         \hline
         $\checkmark$ & $\times$& $\times$ & 72.2 & 69.2\\
         $\checkmark$ & $\checkmark$& $\times$ & 85.0 & 84.3\\
         $\checkmark$ & $\checkmark$& $\checkmark$ & 85.8 & 84.7\\
        \bottomrule
    \end{tabular}}
    \caption{\label{tab:parameter_result} The detailed performance of ModCR with different training strategies. ``MappNet'' indicates the two types of mapping networks. ``$\checkmark$'' represents parameters of the module will be updated during training. The top 3 lines show the experimental results on the VCR~(QR$\rightarrow$A), and the bottom 3 lines is PMR.}
\end{center}
\end{table}

\subsection{Ablation Studies}

\begin{figure*}[t]
    \centering
    \includegraphics[height=0.37\textwidth]{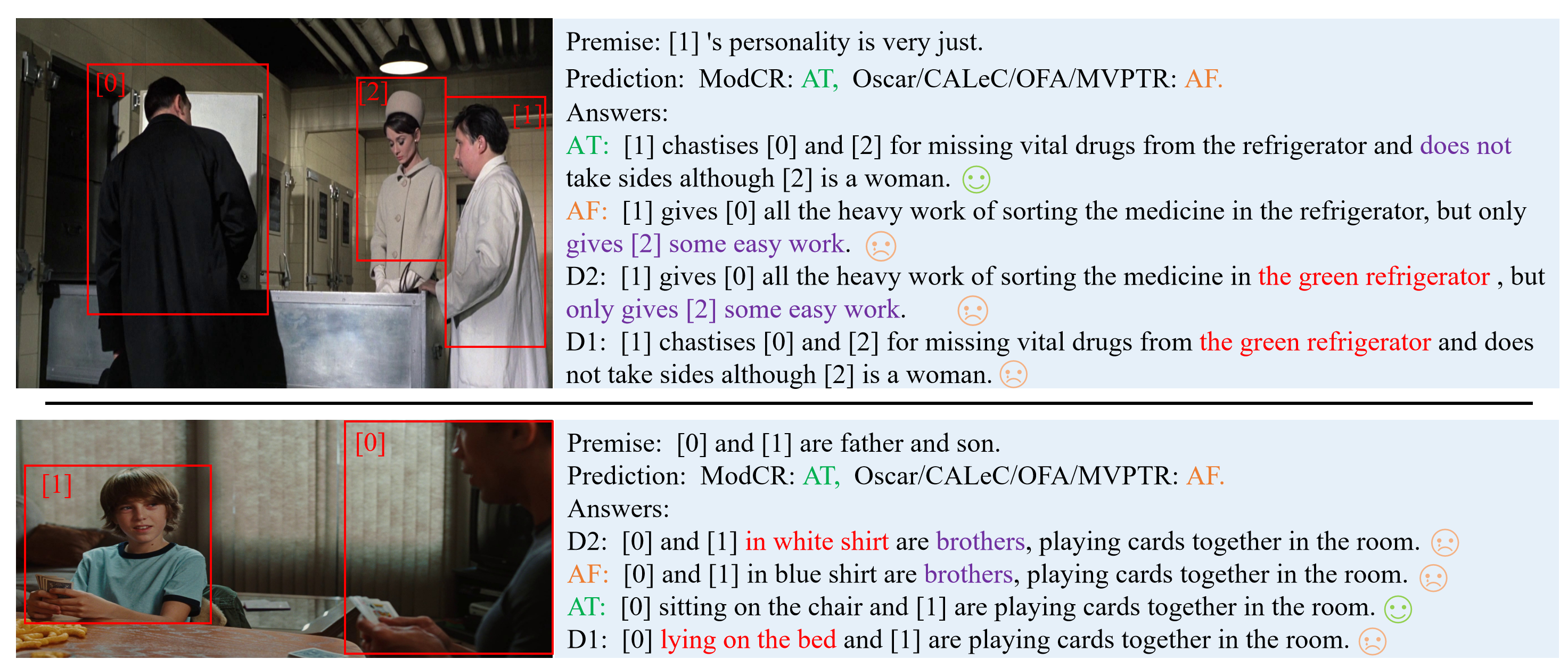}
    \caption{Two cases from the test set of PMR. Different persons are represented by red squares and numbers in the left images. For answer candidates, the red words indicate the content that does not meet the image; the purple indicates the content that is conflict with the premise text clues; and the green and orange indicate the correct option and the wrong answer respectively.}
    \label{fig:cases}
\end{figure*}

To analyze the effectiveness of ModCR in detail, we design multiple model variants and the experimental results are shown in Tables~\ref{tab:module_result} and~\ref{tab:parameter_result}. We select the high-quality PMR (manual annotation and inspection) data set as the experimental scene of ablation studies. For PromptFuse~\citep{liang-etal-2022-modular}, we adopt RoBERTa-L as the backbone and all parameters are updated during training. 

\noindent\textbf{Is Alignment Mapping Network Effective?} From Table~\ref{tab:module_result}, comparing ModCR performances with LA=0 and LA >= 1, we observe that the performance of ModCR drops markedly when it abandons vision-language semantic alignment information. Compared to PromptFuse that randomly initializes two learnable alignment prefix vectors, the proposed alignment mapping network equipped with the multi-grained cross-modal alignmenter is more effective, e.g., PromptFuse vs. RoBERTa-L: 76.5 vs. 75.0, and performance comparisons of ModCR vs. RoBERTa-L. 

\noindent\textbf{Effect of Prefix Length on Model Performance.} From the performance of the visual prefix and alignment prefix at different lengths in Table~\ref{tab:module_result}, we can see that the performance of ModCR varies greatly under different lengths for the two types of prefix. The ModCR performs best when both prefixes are taken as 5. Furthermore, excessively long visual prefixes impair the overall performance, which may be attributed to the fact that redundant and inaccurate visual prefix has an inferior effect on the context learning capability of language model.

\noindent\textbf{Model Performance with Different Training Strategies.}
We present the detailed performance of ModCR with different training strategies on Table~\ref{tab:parameter_result}. By comparing the experimental results of ``frozen VLM'' and ``fine-tune VLM'' on two data sets, we observe that the performance of the proposed method is further improved when all parameters of ModCR are updated during training. Although the training speed is slower, this could further integrate the complementary reasoning capabilities of VLM and LM. In addition, only finetuning MappNet has inferior performances, which may be addressed via pretraining on external large-scale image-text corpus.

\subsection{Case Study}

We report two cases in Figure~\ref{fig:cases} to analyse the performance of models in detail. The premise texts of two samples are about the character (top case) and relationship (bottom one) of persons respectively. Although pre-trained VLMs can infer whether the answer candidate satisfies the image content, they cannot effectively use the premise information to perform reasoning. Contrastly, ModCR utilizes the two-modal semantic information to determine the correct answer. It indicates that regrading two different cues as pre-context states and employing the context reasoning ability of language models is a simple and effective approach for cross modal reasoning tasks. In addition, ModCR could infer the description ``in white shirt'' and ``lying on the bed'' do not meet the image content (the boy wearing blue shirt and sitting on the chair), which may be attributed to the semantic alignmenter. To conclude, the alignment prefix can improve the whole performance of allowing the language model to understand the visual information and perform reasoning.

\section{Conclusion and Future Work}
In this paper, we propose a multi-modal context reasoning approach named ModCR for the scenario of conditional inference on joint visual and textual clues. It regards the given image and text as the two types of pre-context states and infuses them into the language model via the instruction learning method to perform such multi-modal reasoning. The experimental results on two data sets show the effectiveness of ModCR.
For the future, we will explore two research directions: 1) how to improve the context learning capability of pretrained VLMs. 2) exploring the conditional inference on complex visual and textual clues, where it contains multiple clues lying in more modalities.

\section*{Limitations}
The proposed method has several limitations:
1) The current approach achieves hunky context reasoning performance in the cross-modal scene of a single text clue and image, but the context reasoning capability in the scene containing multiple textual and visual clues still needs to be further explored, such as video and long text. 2) From the experimental results, we observed that the visual prefix length greatly impacts the stability of language models infused with visual information. Hence, we still need to explore effective and stable vision-aided language models for natural language processing and multi-modal scenarios. 3) We also hope this work could spark further research on improving the long context reasoning capability of pretrained vision-language models.


\bibliography{custom}
\bibliographystyle{acl_natbib}




\end{document}